\documentclass[letterpaper, 10 pt, conference]{IEEEtran}
\IEEEoverridecommandlockouts

\usepackage[utf8]{inputenc}
\usepackage[T1]{fontenc}
\usepackage{amsmath,amssymb}
\usepackage{graphicx}
\usepackage{booktabs}
\usepackage{cite}
\usepackage{siunitx}
\usepackage{hyperref}
\usepackage{xcolor}
\usepackage{caption}
\usepackage{subcaption}
\usepackage{algorithm}
\usepackage{algpseudocode}
\usepackage{multirow}
\usepackage{listings}

\hypersetup{
    colorlinks=true,
    linkcolor=blue,
    filecolor=magenta,
    urlcolor=cyan,
    citecolor=blue,
}

\title{\LARGE \bf
World-Model-Grounded LLM Planning for AUV and ASV\\
Navigation Near Offshore Wind Farms
}

\author{Markus Buchholz$^{1}$, Ignacio Carlucho$^{2}$ and Yvan R. Petillot$^{2}$%
\thanks{$^{1}$Norwegian Defence Research Establishment (FFI), Kjeller, Norway
        {\tt\small markus.buchholz@ffi.no}}%
\thanks{$^{2}$School of Engineering \& Physical Sciences, Heriot-Watt University, Edinburgh, UK
        {\tt\small \{ignacio.carlucho, y.r.petillot\}@hw.ac.uk}}%
}

\begin{document}

\maketitle
\thispagestyle{empty}
\pagestyle{empty}
\newcommand{\TODO}[1]{\textcolor{red}{\textbf{TODO:} #1}}
% ============================================================================
\begin{abstract}
Large language models can turn a natural-language mission into a sequence of robot actions, but they do not have a sense of physics: they cannot judge how long a command should run, or whether it will make the robot drift into an obstacle. 
We proposed the use of a world model to expand the capabilities of Large Language model-based planners. Our method has three components: a physics-grounded neural world model, a three-phase gradient-based trajectory optimizer, and a Model Predictive Controller (MPC)-style closed-loop replanner with a trust-region guard. The language model decides \emph{what} to do, and the world model decides \emph{how long}, whether that means driving eight thrusters through 6 DOF or two differential thrusters through 3. We evaluate two marine vehicle classes operating near offshore wind infrastructure: a 6-DOF Autonomous Underwater Vehicle (AUV) and a 3-DOF differential-drive Autonomous Surface Vehicle (ASV). In five benchmark missions per platform, both vehicles reach every goal with zero predicted collisions, and both transfer to GazeboSim under ocean current, waves, and thruster dynamics, remaining collision-free and cutting GazeboSim goal-distance error versus the ungrounded baseline by 70-82\% (ASV) and roughly 93\% (AUV), after a residual fine-tuning pass that separately reduces surrogate rollout Root Mean Square Error (RMSE) by 60\% (AUV) and 69\% (ASV). For the ASV we further demonstrate a Vision language model (VLM)-assisted semantic-mapping pipeline that extracts obstacles and environmental context from satellite imagery, nautical charts, and forecast Application Programming Interface (API) instead of onboard sensors, reaching 96\% navigability accuracy as a drop-in replacement for hand-specified obstacle geometry.
\end{abstract}

% ============================================================================
\section{Introduction}
\label{sec:intro}

\begin{figure*}[t]
    \centering

    % \begin{subfigure}[t]{0.40\textwidth}
    %     \centering
    %     \includegraphics[
    %         width=\linewidth,
    %         height=0.34\textheight,
    %         keepaspectratio
    %     ]{figures/fig_1_auv.png}
    %     \caption{AUV}
    %     \label{fig:intuition_a}
    % \end{subfigure}
    % \hfill
    % \begin{subfigure}[t]{0.6\textwidth}
    %     \centering
    %     \includegraphics[
    %         width=\linewidth,
    %         height=0.34\textheight,
    %         keepaspectratio
    %     ]{figures/fig_1_asv.png}
    %     \caption{ASV}
    %     \label{fig:intuition_b}
    % \end{subfigure}
     \includegraphics[
            width=\linewidth,
            height=0.25\textheight,
            keepaspectratio
        ]{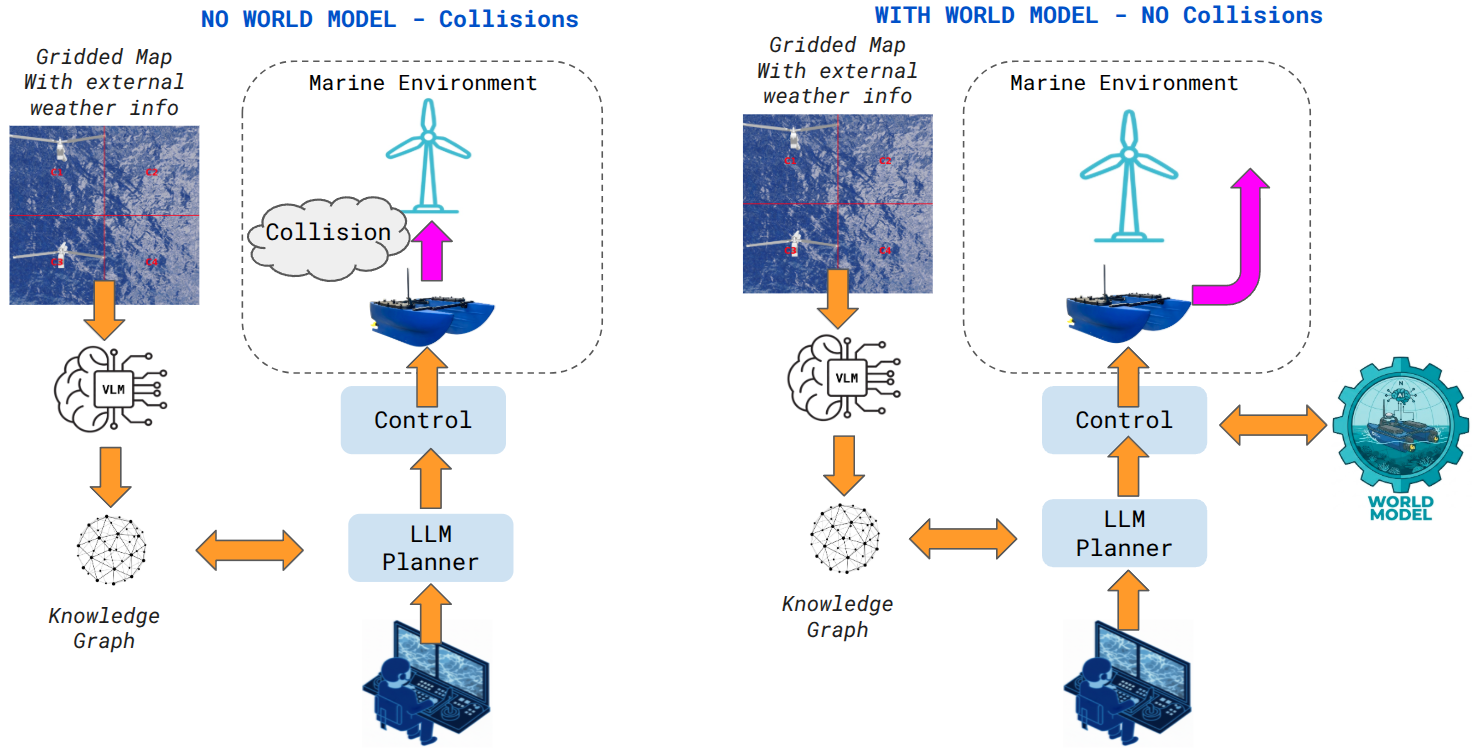}
    %\caption{Why an LLM planner needs a world model, for both vehicle classes. \textbf{(a) AUV:} the LLM translates the mission into thruster commands with no notion of hydrodynamics or drift (left) and collides with the subsea structure; grounding the same plan in the learned world model (right) yields physically feasible step durations and a collision-free approach. \textbf{(b) ASV:} the same contrast for the Blueboat, whose inputs additionally include a gridded map analyzed cell-by-cell by a vision-language model and enriched with forecast wind/current/wave data (Section~\ref{sec:vlm_pipeline}); without grounding (left) the ASV drifts into the turbine tower, while the grounded plan (right) reaches it safely.}
    % \caption{Why an LLM planner needs a world model. The LLM translates the mission into thruster commands with no notion of hydrodynamics or drift (left) and collides with the subsea structure; grounding the same plan in the learned world model (right) yields physically feasible step durations and a collision-free approach. Additionally, the LLM planner can include a gridded map analyzed cell-by-cell by a vision-language model and enriched with forecast wind/current/wave data (Section~\ref{sec:vlm_pipeline}); providing more safe and accurate plans.}

    \caption{Why an LLM planner needs a world model (ASV variant shown; the same contrast holds for the AUV, without the VLM stage). Without grounding (left), the LLM's plan drifts into the turbine tower; with the world model (right), the same plan receives physically feasible step durations and reaches the goal collision-free. The ASV's inputs additionally include a gridded map classified by a vision-language model and enriched with forecast wind/current/wave data (Section~\ref{sec:vlm_pipeline}); the AUV instead uses pre-dive obstacle boxes (Section~\ref{sec:background}), with no VLM stage.
    }
    \label{fig:intuition}
\end{figure*}

Large language models (LLM) have opened a compelling way to plan robot motion: describe a mission in natural language and let the LLM decompose it into executable actions~\cite{huang2022language, ahn2022saycan, buchholz2025urosa}. This works well in manipulation and navigation, where commonsense reasoning about objects and space produces useful plans. Marine robotics near offshore wind infrastructure exposes a fundamental limitation of this approach, for both ASVs and AUVs. A command that looks geometrically reasonable, such as ``move forward, then turn to avoid the turbine,'' can fail catastrophically for reasons the LLM cannot anticipate: thruster lag slows the first seconds of any move, hydrodynamic coupling makes combined surge and sway (or surge and yaw) unpredictable, and current or wind accumulate lateral drift over a mission. As LeCun argues~\cite{lecun2022path}, predicting the next token is the wrong objective for systems that must reason about the physical world, a language model that has processed millions of descriptions of marine vehicles still has no \emph{representation} of what drag or thruster saturation does to a trajectory. It only knows how people \emph{write} about these effects. LeCun's proposed alternative, JEPA~\cite{lecun2022path}, learns compact representations of future world states rather than token sequences; Ha and Schmidhuber~\cite{ha2018world} showed that an agent should \emph{imagine} the consequences of actions in a compact model before committing, and Dreamer~\cite{hafner2020dreamer} proved this works for continuous control by planning entirely within a learned world model before any real action is taken. Fig.~\ref{fig:intuition} captures the consequence for both vehicle classes: the same LLM planner that drives blindly into a subsea structure or a turbine tower reaches the goal cleanly once a world model checks and corrects its plan before execution. The LLM decides \emph{what} to do in both cases, regardless of whether the vehicle has 6 DOF and eight thrusters or 3 DOF and two differential thrusters, and the world model decides \emph{how long}, i.e., whether the plan will survive contact with the physics.

% \begin{figure*}[t]
% \centering
% \includegraphics[width=0.72\textwidth, height=0.40\textheight, keepaspectratio]{figures/fig1_merged_whyWM.png}
% \caption{Why an LLM planner needs a world model, for both vehicle classes. \textbf{Top (AUV):} the LLM translates the mission into thruster commands with no notion of hydrodynamics or drift (left) and collides with the subsea structure; grounding the same plan in the learned world model (right) yields physically feasible step durations and a collision-free approach. \textbf{Bottom (ASV):} the same contrast for the Blueboat, whose inputs additionally include a gridded map analyzed cell-by-cell by a vision-language model and enriched with forecast wind/current/wave data (Section~\ref{sec:vlm_pipeline}); without grounding (left) the ASV drifts into the turbine tower, while the grounded plan (right) reaches it safely. Both rows show the same qualitative plan topology surviving contact with real dynamics only once it is checked before execution.}
% \label{fig:intuition}
% \end{figure*}

Two features of the problem are genuinely vehicle-specific. First, maneuverability: the AUV can move directly sideways to correct a slightly-off plan, while the ASV, like most surface vessels, is nonholonomic - its LLM must instead lay out an explicit turn-forward-turn sequence, and a wrong sequence cannot be corrected mid-execution (Section IV-C, IV-G). Second, knowing where the obstacles are is not free: our AUV missions assume obstacle positions from a pre-dive structural survey, but equipping a low-cost ASV with real-time sonar or LiDAR for complex coastal bathymetry is prohibitively expensive. For the ASV, we therefore also develop a vision-language model (VLM) pipeline that extracts obstacle and environmental context from satellite imagery, nautical charts, and forecast APIs - sensor-cheap data already available for Norwegian and North Sea coastal waters - in place of onboard perception.

In summary, this work contributes: \textbf{(1)} a physics-grounded neural surrogate - an analytical Fossen simulator plus a learned residual MLP - instantiated for both a 6-DOF AUV and a 3-DOF differential-drive ASV with sub-centimeter (AUV) and sub-\SI{0.4}{m} (ASV) open-loop accuracy over \SI{60}{s}; \textbf{(2)} a shared three-phase gradient-based trajectory optimizer and MPC-style replanner with a trust-region guard, unchanged in structure across both vehicles, for which an ablation on the AUV's tightest slalom mission shows the guard is decisive (100\% vs.\ 10\% goal-reach without it); \textbf{(3)} a characterization of the skeleton-topology failure mode common to both platforms, including the ASV's differential-drive no-strafe constraint enforced at the LLM prompt level; \textbf{(4)} a VLM-assisted semantic-mapping pipeline for the ASV that extracts obstacles and environmental context from satellite imagery, nautical charts, and forecast data, validated as a drop-in replacement for hand-specified obstacle geometry; and \textbf{(5)} empirical evaluation on five missions per platform (3-D for the AUV, 2-D for the ASV) around offshore wind infrastructure, with transfer to GazeboSim under ocean currents, waves, and thruster dynamics.
% The contributions of this work are fivefold:
% \begin{enumerate}
%     \item A \textbf{physics-grounded neural surrogate architecture} - an analytical Fossen simulator plus a learned residual MLP - instantiated for both a 6-DOF AUV and a 3-DOF differential-drive ASV, validated with sub-centimeter (AUV) and sub-\SI{0.4}{m} (ASV) open-loop accuracy over \SI{60}{s}.
%     \item A \textbf{shared three-phase gradient-based trajectory optimizer and MPC-style replanner} with a trust-region guard, applied with an unchanged structure to both vehicles; an ablation on the AUV's tightest slalom mission shows the guard is decisive (100\% vs.\ 10\% goal-reach without it).
%     \item A \textbf{characterization of the skeleton-topology failure mode} common to both platforms, including the ASV's differential-drive no-strafe constraint enforced at the LLM prompt level.
%     \item A \textbf{VLM-assisted semantic-mapping pipeline} (ASV) that extracts obstacles and environmental context from satellite imagery, nautical charts, and forecast data, validated as a drop-in replacement for hand-specified obstacle geometry.
%     \item \textbf{Empirical evaluation} on five missions per platform (3-D for the AUV, 2-D for the ASV) around offshore wind infrastructure, with transfer to a higher-fidelity simulator (GazeboSim) under ocean currents, waves, and thruster dynamics.
% \end{enumerate}

The LLM is called once per mission; every subsequent duration-optimization and replanning step runs without querying it again.

% ============================================================================
\section{Related Work}

\textbf{World models and LLM planning.}
Ha and Schmidhuber~\cite{ha2018world} showed that an agent should imagine consequences in a compact model before acting; Dreamer~\cite{hafner2020dreamer} proved this works for continuous control by planning entirely within a learned world model. LeCun~\cite{lecun2022path} formalized the critique of token prediction for physical reasoning, arguing for architectures that predict \emph{representations} of future states - our surrogate does exactly this for both vehicles. On the language side, SayCan~\cite{ahn2022saycan} grounds LLM outputs in affordances, Code as Policies~\cite{liang2023code} generates executable programs, and Inner Monologue~\cite{huang2022inner} and ELLMER~\cite{monwilliams2025ellmer} add feedback to the LLM loop; none insert a differentiable physics model between the LLM and the robot.

\textbf{Physics-informed learning for marine vehicles.}
Neural surrogates exist as black-box models~\cite{hessel2021ship,kim2022uuv} and for thrust allocation~\cite{engelhard2023thrust}, discarding the analytical model entirely; hybrid first-principles-plus-residual models~\cite{greydanus2019hamiltonian,chen2018neural} retain physical structure, which our surrogate follows for both vehicles. High-fidelity simulators including Stonefish~\cite{ciesielski2019stonefish,grimaldi2025stonefish} and
GazeboSim (the latter driven through ArduPilot, SITL and a ROS~2 bridge to replicate the real autopilot stack~\cite{buchholz2025tethered})
% GazeboSim ~\cite{buchholz2025tethered}
serve as digital twins; our surrogate plays a complementary role as a fast differentiable approximation for offline plan search. Broader agentic architectures for marine autonomy have also been explored~\cite{buchholz2025aura,buchholz2025urosa}.

\textbf{Semantic mapping for sensor-cheap navigation (ASV).}
Grid-based semantic mapping from satellite imagery has been demonstrated by OVerSeeC~\cite{rana2026overseec}, and open-vocabulary scene graphs such as ConceptGraphs~\cite{gu2024conceptgraphs} support spatial retrieval for planning; our knowledge graph follows this design but merges cells directly into the differentiable collision field. How well LLMs cope with geospatial context is an open question: the Mil-SCORE benchmark shows current models struggle to combine maps, orders, and reports at scenario scale~\cite{palnitkar2026milscore}, which is why we hand the LLM a short, pre-digested terrain summary rather than raw map data.

% ============================================================================
\section{Background: Unified Marine Vehicle Dynamics}
\label{sec:background}

We use the standard Fossen~\cite{fossen2011} notation throughout. In general, a marine vehicle's kinematics follow $\dot{\boldsymbol{\eta}} = \mathbf{R}(\boldsymbol{\eta})\boldsymbol{\nu}$, relating body-frame velocity $\boldsymbol{\nu}$ to the rate of change of world-frame pose $\boldsymbol{\eta}$ through the rotation (and, in 6-DOF, transformation) matrix $\mathbf{R}$. The kinetics follow
\begin{equation}
    \mathbf{M}\dot{\boldsymbol{\nu}} + \mathbf{C}(\boldsymbol{\nu})\boldsymbol{\nu} + \mathbf{D}(\boldsymbol{\nu}_r)\boldsymbol{\nu}_r + \mathbf{g}(\boldsymbol{\eta}) = \boldsymbol{\tau},
    \label{eq:kinetics}
\end{equation}
where $\mathbf{M}$ is the generalized mass (rigid-body plus added mass), $\mathbf{C}$ is the Coriolis-centripetal matrix, $\mathbf{D}$ is hydrodynamic damping, $\mathbf{g}$ is the restoring vector, $\boldsymbol{\nu}_r=\boldsymbol{\nu}-\boldsymbol{\nu}_c$ is velocity relative to ambient current (or current-plus-wind), and $\boldsymbol{\tau}$ is the control wrench delivered through a first-order actuator lag with time constant $\tau_a$. This single formulation specializes to each platform only in degrees of freedom, actuation, and coefficients (BlueROV2 Heavy~\cite{wu2018bluerov2,vonBenzon2022bluerov2benchmark}; Blueboat~\cite{buchholz2026blueboatrepo}), summarized in Table~\ref{tab:platform_compare}.

\begin{table}[t]
\centering
\small
\caption{Platform specialization of Eq.~\eqref{eq:kinetics}.}
\label{tab:platform_compare}
\scriptsize
\begin{tabular}{@{}lll@{}}
\toprule
 & AUV (BlueROV2 Heavy) & ASV (Blueboat) \\
\midrule
DOF & 6, $\boldsymbol{\eta}=[x,y,z,\phi,\theta,\psi]$ & 3, $\boldsymbol{\eta}=[x,y,\psi]$ \\
Actuation & 8$\times$T200, full $\mathbf{B}$ & 2$\times$diff.\ thrusters, no-strafe \\
Mass & \SI{13.5}{kg} & \SI{16.0}{kg} \\
Planner state & $(\boldsymbol{\eta},\boldsymbol{\nu})\!\in\!\mathbb{R}^{12}$, SoC separate & $\mathbf{s}\!\in\!\mathbb{R}^{9}$, incl.\ $(n_p,n_s)$, SoC \\
Actuator lag & first-order (validated, Sec.~\ref{sec:hifi}) & first-order, $\tau_a{=}\SI{0.15}{s}$, $c{=}0.78$ \\
\bottomrule
\end{tabular}
\end{table}

What makes the Blueboat's control wrench distinctive is that it is generated by only two collinear thrusters,
\begin{equation}
\boldsymbol{\tau}=\begin{bmatrix} T_p + c\,T_s\\ 0\\ 0.5\,B\,(T_p - c\,T_s)\end{bmatrix},
\label{eq:asv_wrench}
\end{equation}
where $T_p,T_s$ are port/starboard thrust, $B{=}\SI{0.41}{m}$ is the beam, and $c{=}0.78$ the measured starboard efficiency correction. This is the direct analogue of the AUV's full allocation-matrix wrench $\boldsymbol{\tau}=\mathbf{B}\mathbf{f}$, but with no lateral force component available at all, which is the physical source of the no-strafe constraint discussed in Section~\ref{sec:interface}.

% ============================================================================
\section{Method}

\subsection{Surrogate World Model Architecture}
\label{sec:surrogate}

The surrogate predicts the next state as
\begin{equation}
    s_{t+1} = s_t + \Delta_{\text{physics}}(s_t, a_t) + \Delta_{\text{residual}}(s_t, a_t),
    \label{eq:residual}
\end{equation}
where $\Delta_{\text{physics}}$ is an analytical RK4 step from the Fossen simulator of Section~\ref{sec:background} and $\Delta_{\text{residual}}$ is a neural-network correction - the physically-grounded world model Ha \& Schmidhuber~\cite{ha2018world} and LeCun~\cite{lecun2022path} argue for, instantiated in each vehicle's own state space. Three advantages follow from the hybrid design, for both platforms: the analytical half enforces physical limits that must always hold - angles wrap correctly, actuators saturate at their real limits, and battery charge only decreases; the network only corrects what the analytical model misses, requiring far less training data than an end-to-end model; and a large residual at an operating point signals where the analytical model is inaccurate. Obstacle geometry is never encoded in the surrogate; it enters only through the cost function (Section~\ref{sec:gradient_opt}), so the same world model serves any mission without retraining.

The residual network is a four-layer MLP with hidden layers [512, 512, 256, 128], GELU activations, layer normalization, and dropout 0.05 for \emph{both} vehicles; the two instantiations differ only in input/output dimensionality (Table~\ref{tab:surrogate_dims}).

\begin{table}[t]
\centering
\small
\caption{Surrogate input/output dimensionality per platform.}
\label{tab:surrogate_dims}
\scriptsize
\begin{tabular}{@{}l p{0.37\columnwidth} p{0.37\columnwidth}@{}}
\toprule
 & AUV & ASV \\
\midrule
Input dim.\  & 50 (state, action, $\sin$/$\cos$ angles, $dt$) & 22 (norm.\ state, $\sin$/$\cos$ yaw, norm.\ action/delta) \\
Output dim.\ & 19-D state delta & 9-D state delta (trainable zero-point) \\
\bottomrule
\end{tabular}
\end{table}

\subsection{Training}
\label{sec:training}

Both surrogates use the same composite loss: a one-step prediction term, an absolute next-state term, an angle-aware term using sine/cosine features, a 50-step rollout term that forces long-horizon accuracy, an equilibrium term that removes trim-point drift, and a residual-magnitude regularizer that keeps the neural correction small and interpretable - optimized with Adam (lr~$10^{-3}$, batch~512, early stopping). Both training sets are generated from random smooth/step commands, stabilizing-controller rollouts, and macro-action sequences, with 25\% of samples near equilibrium to ensure accurate passive-stabilization prediction. The AUV model trains on 500\,k transitions and converges in 48 epochs (validation loss $\sim10^{-4}$; final RMSE below \SI{0.06}{mm} position, \SI{0.3}{mm/s} velocity); the ASV model trains on 200\,k transitions (loss weights $w_1{=}1.0,w_2{=}0.5,w_3{=}2.0,w_4{=}0.5,w_{\text{eq}}{=}1.0,w_{\text{res}}{=}0.1$; patience 15).

\subsection{LLM Planning Interface}
\label{sec:interface}

% Both systems use \texttt{gemma4:26b}, served locally via Ollama at temperature 0.3, expose the LLM a fixed macro-action vocabulary, and invoke it exactly once per mission: the LLM returns an ordered list of \texttt{(macro, duration)} pairs with no numeric durations attached, and the differentiable world model finds how long each step should run. The \textbf{AUV} interface exposes nine macros - \texttt{stabilise, move\_forward/backward, sway\_left/right, ascend, descend, turn\_left/right} - covering full 3-D motion including direct lateral strafing. The \textbf{ASV} interface exposes five macros - \texttt{stabilize, move\_forward/backward, turn\_left/right} - with no sway macro at all: the Blueboat's two collinear thrusters (Eq.~\eqref{eq:asv_wrench}) cannot generate a lateral force component, so the LLM must reason in turn-forward-turn sequences rather than direct lateral detours, and the no-strafe constraint is additionally stated explicitly in the prompt. Both interfaces score a candidate plan with the same cost.
Both systems use \texttt{gemma4:26b}, served locally via Ollama at temperature 0.3, expose the LLM to a fixed macro-action vocabulary, and invoke it exactly once per mission: the LLM returns an ordered list of \texttt{(macro, duration)} pairs with no numeric durations attached, and the differentiable world model finds how long each step should run. The \textbf{AUV} interface exposes nine macros - \texttt{stabilize, move\_forward/backward, sway\_left/right, ascend, descend, turn\_left/right} - covering full 3-D motion including direct lateral strafing. The \textbf{ASV} interface exposes five macros - \texttt{stabilize, move\_forward/backward, turn\_left/right} - with no sway macro at all: the Blueboat's two collinear thrusters (Eq.~\eqref{eq:asv_wrench}) cannot generate a lateral force component, so the LLM must reason in turn-forward-turn sequences rather than direct lateral detours, and the no-strafe constraint is additionally stated explicitly in the prompt. Both interfaces score a candidate plan with the same cost, which the gradient optimizer of Section~\ref{sec:gradient_opt} minimizes over the per-step durations proposed by the LLM:
%\TODO{we need to explain who is optimizing this cost}
\begin{equation}
    J = d_{\text{goal}} + 0.1\,L_{\text{path}} + 100\!\sum_t \phi_{\text{coll}}(\mathbf{p}_t),
    \label{eq:cost}
\end{equation}
where $d_{\text{goal}}$ is final distance to goal, $L_{\text{path}}$ is path length, and $\phi_{\text{coll}}$ is the smooth obstacle-membership field of Eq.~\eqref{eq:smooth_coll}.

\subsection{Gradient-Based Trajectory Optimization}
\label{sec:gradient_opt}
This step belongs to the gradient optimizer, not the LLM: taking the LLM's macro-action skeleton as fixed, it searches over the per-step durations to minimize Eq.~\eqref{eq:cost}. Because the residual network $\Delta_{\text{residual}}$ is differentiable, this search can follow gradients directly through the surrogate, while the frozen analytical component $\Delta_{\text{physics}}$ is treated as a constant at each step. Obstacles enter through a smooth sigmoid repulsion field
\begin{equation}
    \phi_{\text{coll}}(\mathbf{p}) = \sum_{o \in \mathcal{O}} \prod_{k \in \{x,y[,z]\}} \sigma\!\left(\alpha\bigl(h_k^{(o)} - |p_k - c_k^{(o)}|\bigr)\right),
    \label{eq:smooth_coll}
\end{equation}
where $c_k^{(o)}, h_k^{(o)}$ are the center and half-extent of obstacle $o$ along axis $k$ and $\alpha{=}10$ controls boundary sharpness; the AUV case takes the product over $k\in\{x,y,z\}$, while the ASV's planar workspace drops the $z$-term and takes the product over $\{x,y\}$ only. Optimization proceeds in three phases, identical in structure for both vehicles (Algorithm~\ref{alg:planner}): \emph{Phase~1} evaluates ten uniform duration scales $\{0.5\text{-}5.0\times\}$ and keeps the best; \emph{Phase~2} refines individual step durations by adaptive coordinate descent with deltas $\{\pm2.0,\pm1.0,\pm0.5,\pm0.2,\pm0.05\}$; \emph{Phase~3} runs a final 15-iteration Adam pass jointly optimizing all durations through the differentiable surrogate, exploiting gradients near obstacle boundaries.

\begin{algorithm}[t]
\caption{Gradient-Optimized World-Model Planning}
\label{alg:planner}
\begin{algorithmic}[1]
\Require Mission $M$, LLM $C$, world model $W$, candidates $K$
\State $\Pi \gets$ parse($C$.generate(build\_prompt($M$, $K$)))
\State $\Pi \gets \Pi \cup$ geometric\_candidates($M$)
\For{each skeleton $\pi$ in $\Pi$}
    \State \textbf{Phase 1:} $s^* \gets \arg\min_{s \in \mathcal{S}} J(W.\text{sim}(M, \text{scale}(\pi, s)))$
    \State \textbf{Phase 2:} adaptive coordinate descent ($\Delta \in \{2.0 \ldots 0.05\}$)
    \State \textbf{Phase 3:} Adam gradient pass (differentiable $W$)
    \State $J_\pi \gets$ best cost from phases 1-3
\EndFor
\State \Return $\arg\min_\pi J_\pi$
\end{algorithmic}
\end{algorithm}

\subsection{MPC-Style Closed-Loop Replanning}
\label{sec:mpc}

Open-loop execution is sensitive to modeling error and unmodeled disturbance. Both systems wrap the gradient planner in an MPC-style loop: after each executed macro step, the measured state (Doppler Velocity Log, DVL, for the AUV; GPS for the ASV) is fed back and the remaining skeleton is re-optimized from that state, with no further LLM queries. Each replan is accepted only under a \emph{trust-region guard}: if the re-optimized remaining duration collapses below half the warm-start total, the replan is rejected and the previous remaining plan is kept, guarding against the surrogate mis-predicting goal-reach when queried from a drifted, off-distribution state. This guard is the decisive factor on the AUV's tightest slalom mission in GazeboSim: without it, duration collapse drops goal-reach from 100\% to 10\% (Section~\ref{sec:experiments}). The ASV extends the same guard with an environment-triggered threshold: when the VLM pipeline (Section~\ref{sec:vlm_pipeline}) reports significant wave height ($H_s>\SI{1.0}{m}$), the collapse threshold is raised from $0.5\times$ to $0.7\times$, making the replanner more conservative in high sea states.

\subsection{VLM-Assisted Semantic Mapping (ASV)}
\label{sec:vlm_pipeline}
The benchmark missions above use hand-specified bounding boxes from wind-farm engineering drawings. Real coastal navigation (fjord entrances, islands, shallow sills, submerged rocks) cannot be pre-specified this way, and equipping a low-cost Blueboat with sonar or LiDAR sufficient to detect them in real time is prohibitively expensive and power-intensive. The data to replace those sensors already exists: Kartverket publishes nautical charts and satellite imagery of the Norwegian coast~\cite{kartverket}, yr.no provides wind forecasts, and Copernicus Marine Service provides current and wave state~\cite{copernicusmarine}. Our pipeline turns these sources into planner-ready obstacles without touching the planner or gradient optimizer: a top-down map is tiled into \SI{1}{m}$\times$\SI{1}{m} cells (only cells within a corridor of $3\times$ the vehicle beam around candidate routes are queried, roughly 400 per mission); each cell patch is classified by \texttt{gemma4:26b} (the same model that plans) into one of six terrain classes (\textit{open water, shallow water, submerged rock, surface obstacle, navigable channel, restricted zone}); and forecast wind/current/wave data is attached to each cell through the same API interface that would query yr.no and Copernicus in the field (answered here by a filter over the simulator configuration, since GazeboSim has no live chart or forecast). A cell record looks like:
\begin{lstlisting}
{"cell_id":"C3","terrain_class":"wind_turbine",
 "wind_turbine":true,"traversability":0.0,
 "confidence":0.94,"wind_ms":8.2,"wind_dir_deg":247,
 "current_ms":0.18,"current_dir_deg":220,
 "wave_height_m":0.6,"tidal_state":"flood"}
\end{lstlisting}
The \texttt{terrain\_class}/\texttt{traversability} fields drive obstacle extraction; the environmental fields feed three consumers: wind direction informs which detour side is cheaper, current speed sets the obstacle inflation $r_{\text{inflate}} = r_{\text{vehicle}}+r_{\text{safety}}+v_c\,t_{\text{step}}$, and wave height above $H_s{=}\SI{1.0}{m}$ triggers the conservative trust-region threshold of Section~\ref{sec:mpc}. Cell records are ingested into a Neo4j knowledge graph following ConceptGraphs~\cite{gu2024conceptgraphs}, and contiguous non-traversable cells are merged into bounding boxes ready for Eq.~\eqref{eq:smooth_coll}; the LLM receives these boxes together with a natural-language terrain/environment summary rather than raw map data, addressing the long-context geospatial-reasoning weakness reported for current LLMs~\cite{palnitkar2026milscore}. On 180 held-out manually labeled cells, it reaches 96\% accuracy on the navigable/non-navigable split and 87\% on the six-class taxonomy, with the 11\% of low-confidence ($<$0.7) cells flagged for review.

\subsection{Skeleton-Topology Failure Mode}
\label{sec:skeleton}

Across 200 planning trials per platform, proposed skeletons were syntactically valid in all 200 AUV trials and 197/200 (98.5\%) ASV trials, the three ASV failures being malformed JSON caught by the parser and replaced by a geometric fallback. \emph{Topology} (as opposed to formatting) was occasionally suboptimal on the AUV: in four cases, the LLM proposed a direct pass toward an obstacle requiring a lateral detour, and the optimizer resolved this silently in all four by stretching the detour timing to zero-collision cost, without restructuring the sequence. On the ASV, the no-strafe constraint was never violated, and no skeleton required topological repair. In neither case can the optimizer recover from a genuinely wrong topological choice; it only adjusts durations within the fixed skeleton, so this remains the shared open failure mode (Section~\ref{sec:conclusion}).

\begin{figure*}[t]
\centering
\includegraphics[width=0.70\textwidth, height=0.19\textheight, keepaspectratio]{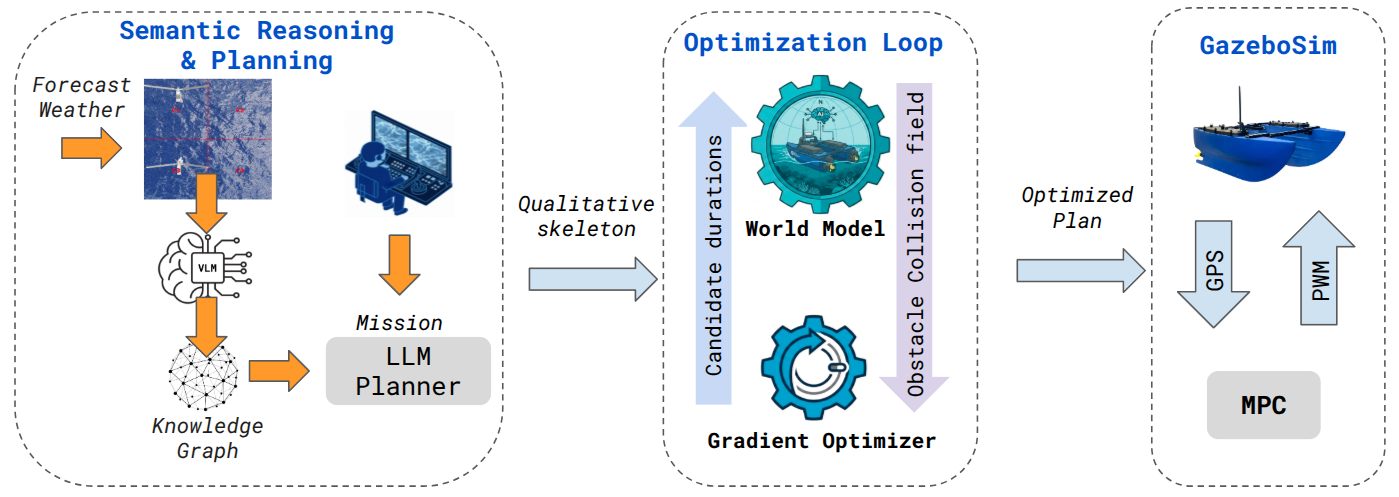}
% \caption{System architecture (ASV variant shown; the AUV variant is structurally identical). \textbf{Planning:} for the ASV, a satellite/chart image is tiled and classified by a vision-language model, augmented with live wind/current/wave/tidal data, and stored in a knowledge graph that merges non-traversable cells into obstacle bounding boxes (Section~\ref{sec:vlm_pipeline}); for the AUV, obstacle boxes come directly from pre-dive structural drawings instead. The LLM is invoked once, returning a qualitative macro-action skeleton with no timing information. \textbf{Optimization loop:} the gradient optimizer proposes candidate step durations; the differentiable world model rolls out the predicted trajectory against the obstacle collision field and returns the cost, repeating across three phases (global scale search, coordinate descent, Adam) until an optimized, collision-free plan emerges. \textbf{Execution:} the plan is dispatched to GazeboSim, where the vehicle is flown by its real autopilot stack; the AUV replanning loop reads back DVL measurements where the ASV reads GPS, after which the MPC-style cycle re-optimizes the remaining skeleton from the measured state.}
\caption{ System architecture (ASV variant shown; the AUV variant is structurally identical). \textbf{Planning:} for the ASV, a satellite/chart image is tiled and classified by a vision-language model and enriched with live wind/current/wave data, merged into obstacle bounding boxes (Section~\ref{sec:vlm_pipeline}); the AUV uses pre-dive structural drawings instead. The LLM is invoked once, returning a macro-action skeleton with no timing information. \textbf{Optimization loop:} the gradient optimizer proposes step durations; the world model scores each rollout against the obstacle collision field across three phases (scale search, coordinate descent, Adam) until a collision-free plan emerges. \textbf{Execution:} the plan runs on GazeboSim's real autopilot stack; the AUV loop reads DVL where the ASV reads GPS, after which MPC re-optimizes the remaining skeleton from the measured state.}
\label{fig:architecture}
\end{figure*}

% ============================================================================
\section{Experiments}
\label{sec:experiments}

\subsection{Setup}
\label{sec:setup}

Both ASV and AUV are evaluated in two environments: a lightweight, GPU-accelerated analytical simulator implementing the Fossen equations of Section~\ref{sec:background} (surrogate training and the majority of planning experiments), and GazeboSim, driven through ArduPilot, SITL and a MAVLink-to-ROS~2 bridge~\cite{buchholz2025tethered}, extended with marine plugins for current, wind, and waves (single PC: Intel i9-14th gen, RTX~4080, 128\,GB RAM). Five benchmark missions per platform are framed as stages of an offshore wind-farm inspection: the AUV missions are in 3-D around wind-turbine foundations and cable trays (\texttt{straight\_line, diagonal\_descent, avoid\_turbine, slalom, obstacle\_field}); the ASV missions are in 2-D around wind-turbine towers (\texttt{straight\_line, diagonal\_survey, avoid\_tower, slalom, obstacle\_field}). Four planners are compared per platform: \emph{Gradient+MPC} (ours, closed loop), \emph{Gradient} (ours, open loop), \emph{WM+uniform scaling} (ablation), and \emph{No~WM} (baseline, raw LLM plan executed unchecked). All planners use the same LLM and prompt template per platform, so any difference is due to the optimization pipeline alone.

\subsection{Surrogate Self-Consistency and Analytical-Simulator Results}
\label{sec:validation}

% We use \textit{validation} in two distinct senses across this section and the next, and it is worth being explicit about what each one checks. Here, the surrogate (analytical Fossen model plus residual MLP) is compared against \emph{the analytical simulator it was trained to approximate}: this verifies self-consistency between the learned model and its own training target, not agreement with an independent ground truth. Section~\ref{sec:hifi} then tests transfer to GazeboSim, a second, independently implemented simulator with unmodelled dynamics the analytical model does not capture - a stronger, sim-to-sim test, though still not real-hardware data; real-world trials remain future work (Section~\ref{sec:conclusion}).

% Before either surrogate is trusted for planning, its simulator is verified by twelve automated component checks (mass/Coriolis/damping structure, actuator time constant, monotonic battery discharge, drift direction, sensor-noise statistics); all twelve pass for both platforms. Table~\ref{tab:validation_both} summarizes open-loop rollout accuracy: the AUV surrogate holds sub-centimeter position RMSE at \SI{60}{s} (a \SI{120}{s} long-horizon maneuver gives \SI{0.020}{m} error, energy tracking $R^2{=}1.000$); the ASV surrogate holds \SI{0.391}{m} position RMSE at \SI{60}{s}, with a closed-loop station-keeping controller staying within \SI{0.71}{m} of the origin over the same horizon.

We use \textit{validation} in two distinct senses across this section and the next, and it is worth being explicit about what each one checks. Here, the surrogate (analytical Fossen model plus residual MLP) is compared against \emph{the analytical simulator it was trained to approximate}: this verifies self-consistency between the learned model and its own training target, not agreement with an independent ground truth. Section~\ref{sec:hifi} then tests transfer to GazeboSim, a second, independently implemented simulator with unmodelled dynamics the analytical model does not capture - a stronger, sim-to-sim test, though still not real-hardware data; real-world trials remain future work (Section~\ref{sec:conclusion}). Even within this self-consistency check, some residual error is expected rather than a sign of surrogate failure: the surrogate's $\Delta_{\text{physics}}$ term uses a fast integration step suited to online planning, while the trajectories it is checked against are generated at finer resolution, so part of the reported RMSE reflects this training/inference discretization gap rather than an external modeling error. This also explains why the two platforms' self-consistency errors are not directly comparable in scale (Table~\ref{tab:validation_both}): the residual network absorbs a different amount of this gap depending on each platform's dynamics and integration settings.

Before either surrogate is trusted for planning, its simulator passes twelve automated component checks (mass/Coriolis/damping structure, actuator dynamics, sensor-noise statistics) for both platforms. Table~\ref{tab:validation_both} summarizes open-loop rollout accuracy; over a longer \SI{120}{s} maneuver the AUV surrogate gives \SI{0.020}{m} error with energy tracking $R^2{=}1.000$, and a closed-loop station-keeping controller holds the ASV within \SI{0.71}{m} of the origin over \SI{60}{s}.

\begin{table}[t]
\centering
\small
\caption{Open-loop rollout RMSE against each analytical simulator (50 random trajectories per platform).}
\label{tab:validation_both}
\begin{tabular}{@{}lccc@{}}
\toprule
Horizon & AUV Pos.\ [m] & ASV Pos.\ [m] & ASV Yaw [rad] \\
\midrule
\SI{10}{s} & 0.0011 & 0.038 & 0.137 \\
\SI{30}{s} & 0.0029 & 0.172 & 0.331 \\
\SI{60}{s} & 0.0060 & 0.391 & 0.571 \\
\bottomrule
\end{tabular}

\vspace{2pt}
\raggedright\scriptsize
ASV yaw RMSE is reported directly at each rollout horizon in the source validation; the AUV's analytical-simulator validation tracked position and velocity RMSE at each horizon but verified orientation only as a qualitative bound (roll/pitch~$<$~\SI{0.002}{rad} under passive stabilization, Section~\ref{sec:validation}) rather than a rollout-horizon RMSE, so no directly comparable AUV yaw number exists at these three horizons.
\end{table}

Table~\ref{tab:results_both} presents aggregate planning results under an identical evaluation protocol per platform (same four planners, same cost-function form, $N{=}3$ trials/mission). Both world-model-grounded planners reach 100\% of goals with zero collisions on both platforms; the ungrounded baseline reaches 0\% on both, colliding on every mission with obstacles (up to 52 collisions on the AUV's tightest missions, up to 68 on the ASV's). Gradient+MPC achieves \SI{0.18}{m} (AUV) and \SI{0.26}{m} (ASV) mean goal distance; the Gradient variant's \SI{0.21}{m}/\SI{0.28}{m} are, respectively, 58\%/61\% better than uniform scaling and 94\%/96\% better than the ungrounded baseline. Per-step duration optimization earns its keep specifically on obstacle missions on both platforms, because a single global scale cannot assign short durations to a tight lateral (AUV) or turn-forward-turn (ASV) detour and long durations to the straight legs.

\begin{table}[t]
\centering
\small
\setlength{\tabcolsep}{4pt}
\renewcommand{\arraystretch}{0.92}
\caption{Aggregate analytical-simulator planning results, both platforms ($N{=}3$ trials/mission, 5 missions).}
\label{tab:results_both}
\begin{tabular}{@{}llcccc@{}}
\toprule
 & Planner & Coll.-Free & Goal Reach & Dist.\,[m] & Coll. \\
\midrule
\multirow{4}{*}{AUV} & Grad+MPC & \textbf{1.00} & \textbf{1.00} & \textbf{0.18} & \textbf{0.0} \\
 & Gradient & \textbf{1.00} & \textbf{1.00} & 0.21 & \textbf{0.0} \\
 & WM+unif.\ & \textbf{1.00} & \textbf{1.00} & 0.50 & \textbf{0.0} \\
 & No~WM & 0.40 & 0.00 & 3.29 & 27.2 \\
\midrule
\multirow{4}{*}{ASV} & Grad+MPC & \textbf{1.00} & \textbf{1.00} & \textbf{0.26} & \textbf{0.0} \\
 & Gradient & \textbf{1.00} & \textbf{1.00} & 0.28 & \textbf{0.0} \\
 & WM+unif.\ & \textbf{1.00} & \textbf{1.00} & 0.71 & \textbf{0.0} \\
 & No~WM & 0.60 & 0.00 & 7.10 & 24.0 \\
\bottomrule
\end{tabular}
\end{table}

\subsection{GazeboSim Transfer Results}
\label{sec:hifi}

Self-consistency (Section~\ref{sec:validation}) does not test transfer. GazeboSim adds unmodelled effects for both platforms: nonlinear thruster transients (ASV: $\tau_a{=}\SI{0.55}{s}$ vs.\ the analytical \SI{0.15}{s}), time-varying current, and drag coupling, plus wave forcing for the ASV. Fine-tuning only the residual network on \SI{250}{k} GazeboSim transitions (under three minutes on an RTX~4080) reduces \SI{60}{s} open-loop position RMSE by 60\% (AUV, \SI{1.05}{m}$\to$\SI{0.42}{m}) and 69\% (ASV, on the macro-action trajectories the planner actually uses, \SI{1.84}{m}$\to$\SI{0.57}{m}).

Table~\ref{tab:gazebo_both} reports planning results ($N{=}10$ trials/mission-planner pair). The AUV's gradient planner reaches every goal with zero collisions; closed-loop replanning roughly halves the turbine-avoidance goal error (\SI{0.52}{m} vs.\ \SI{0.83}{m} open-loop) and the trust-region guard keeps the slalom mission at 100\% goal-reach by rejecting an off-distribution duration collapse (Section~\ref{sec:mpc}). The ASV's grounded planners stay fully collision-free but reach only 20\% under the \SI{1}{m} threshold; with only $N{=}10$ trials this binary cutoff is high-variance and obscures mean distances that clearly separate the planners (Table~\ref{tab:gazebo_both}). The informative comparison is instead 0.0 vs.\ 4.6 mean collisions and \SI{1.5}{}-\SI{1.6}{m} vs.\ \SI{8.8}{m} mean goal distance - both grounded ASV planners cut goal error by 70-82\% relative to the ungrounded baseline.

\begin{table}[t]
\centering
\small
\setlength{\tabcolsep}{4pt}
\renewcommand{\arraystretch}{0.92}
\caption{Aggregate GazeboSim planning results, both platforms ($N{=}10$ trials/mission-planner; mean$\pm$std distance). AUV figures assembled from per-mission GazeboSim results (Section~\ref{sec:hifi}); ``-'': not separately reported at the per-planner level.}
\label{tab:gazebo_both}
\begin{tabular}{@{}llcccc@{}}
\toprule
 & Planner & Succ.\ & Dist.\ [m] & Coll. \\
\midrule
\multirow{4}{*}{AUV} & Grad+MPC & 1.00 & $0.58{\pm}0.19$ & 0.0 \\
 & Gradient & 1.00 & $0.62{\pm}0.13$ & 0.0 \\
 & WM+unif.\ & 0.00 & $3.17{\pm}0.90$ & - \\
 & No~WM & 0.00 & $8.53{\pm}1.31$ & $\sim$18 \\
\midrule
\multirow{4}{*}{ASV} & Grad+MPC & 0.20 & $1.54{\pm}0.58$ & 0.0 \\
 & Gradient & 0.20 & $1.63{\pm}0.56$ & 0.0 \\
 & WM+unif.\ & 0.20 & $2.55{\pm}1.94$ & 0.0 \\
 & No~WM & 0.00 & $8.79{\pm}10.17$ & 4.6 \\
\bottomrule
\end{tabular}
\end{table}

\begin{figure*}[t]
\centering

%================ LEFT GROUP =====================
\begin{minipage}[t]{0.42\textwidth}
\centering

% Top (centered)
\begin{subfigure}[t]{0.46\linewidth}
    \centering
    \includegraphics[width=\linewidth,height=3.0cm,keepaspectratio]
    {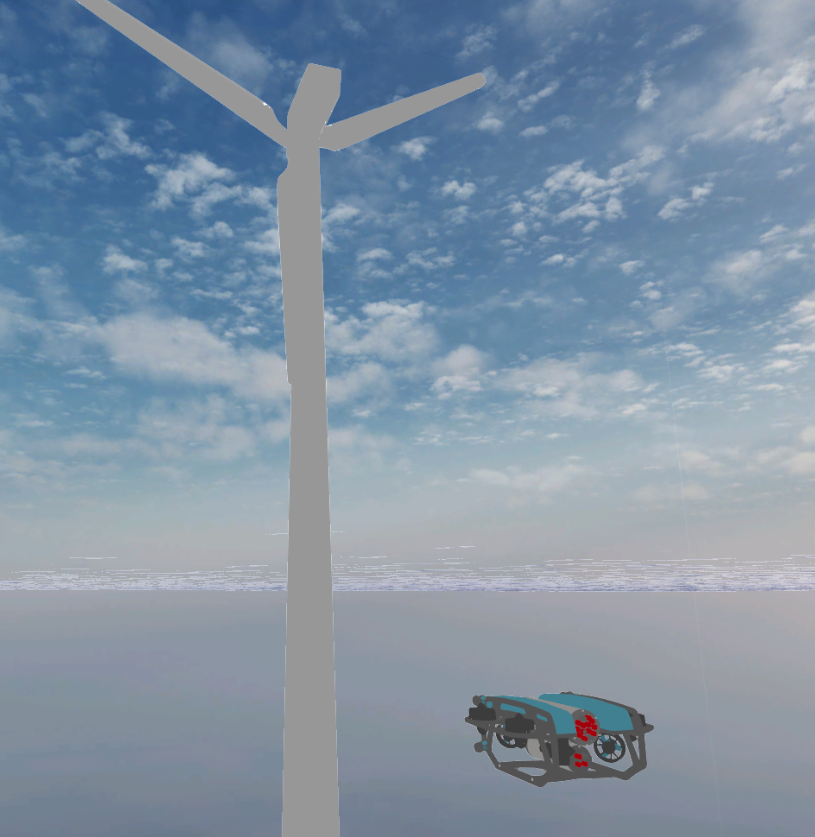}
    \caption{GazeboSim (AUV)}
\end{subfigure}

\vspace{0.6em}

% Bottom row
\begin{subfigure}[t]{0.4\linewidth}
    \centering
    \includegraphics[width=\linewidth,height=4.2cm,keepaspectratio]
    {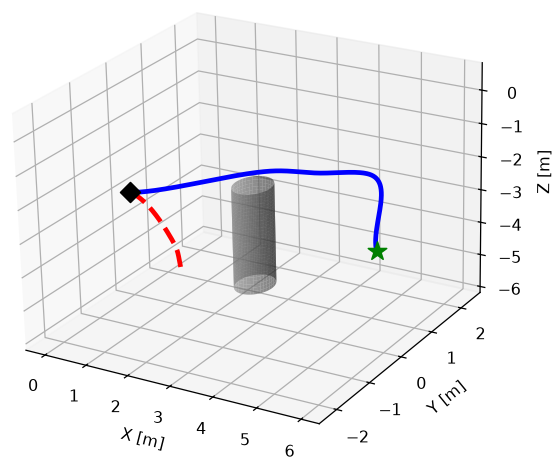}
    \caption{AUV 3-D}
\end{subfigure}
\hfill
\begin{subfigure}[t]{0.5\linewidth}
    \centering
    \includegraphics[width=\linewidth,height=5.2cm,keepaspectratio]
    {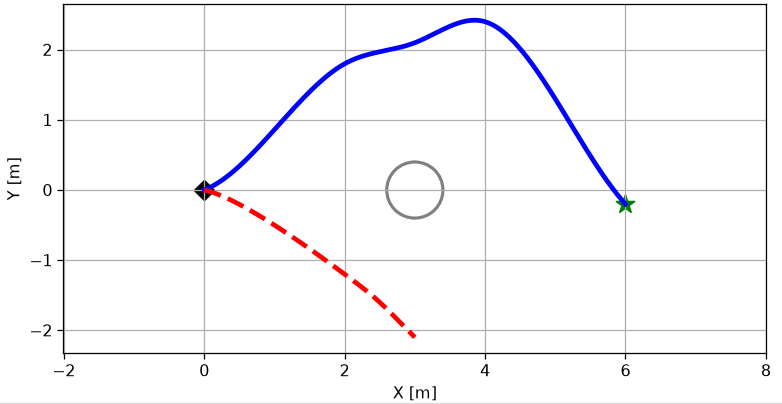}
    \caption{AUV top-down}
\end{subfigure}

\end{minipage}
\hfill
%================ RIGHT GROUP ====================
\begin{minipage}[t]{0.52\textwidth}
\centering

% First row
\begin{subfigure}[t]{0.46\linewidth}
    \centering
    \includegraphics[width=\linewidth,height=3.0cm,keepaspectratio]
    {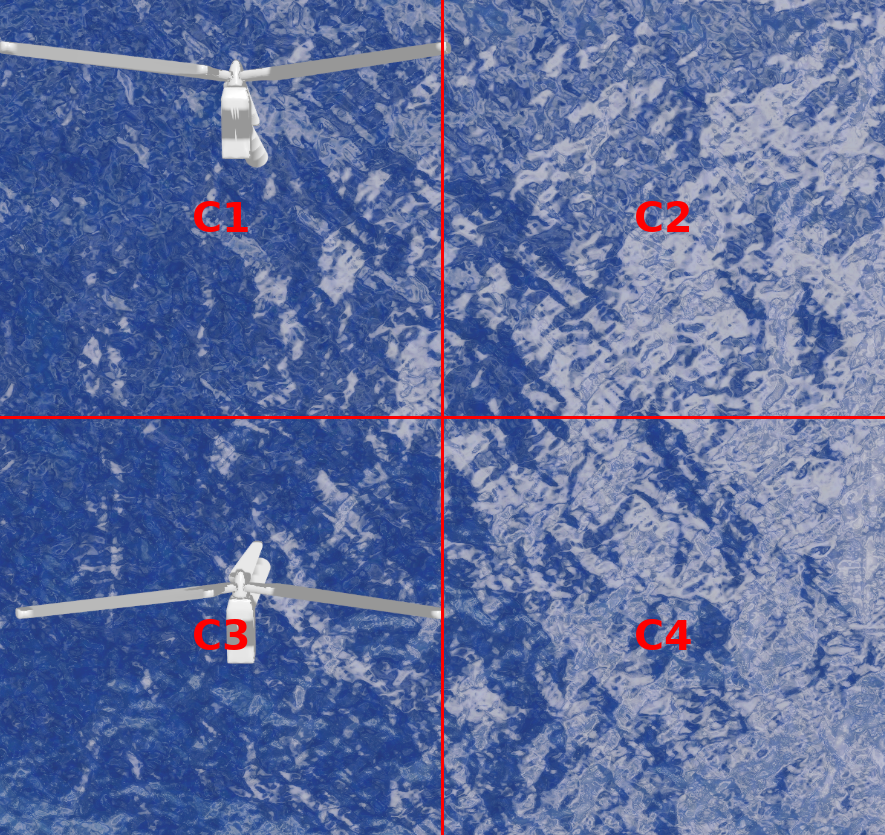}
    \caption{VLM grid cells}
\end{subfigure}
\hfill
\begin{subfigure}[t]{0.46\linewidth}
    \centering
    \includegraphics[width=\linewidth,height=3.0cm,keepaspectratio]
    {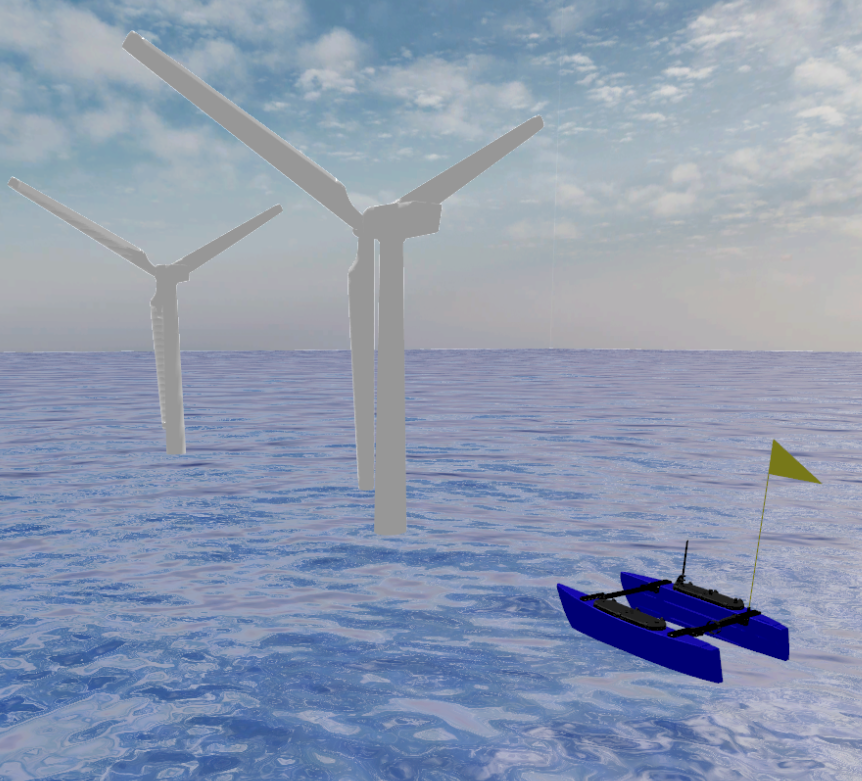}
    \caption{GazeboSim (ASV)}
\end{subfigure}

\vspace{0.6em}

% Second row
\begin{subfigure}[t]{0.46\linewidth}
    \centering
    \includegraphics[width=\linewidth,height=3.0cm,keepaspectratio]
    {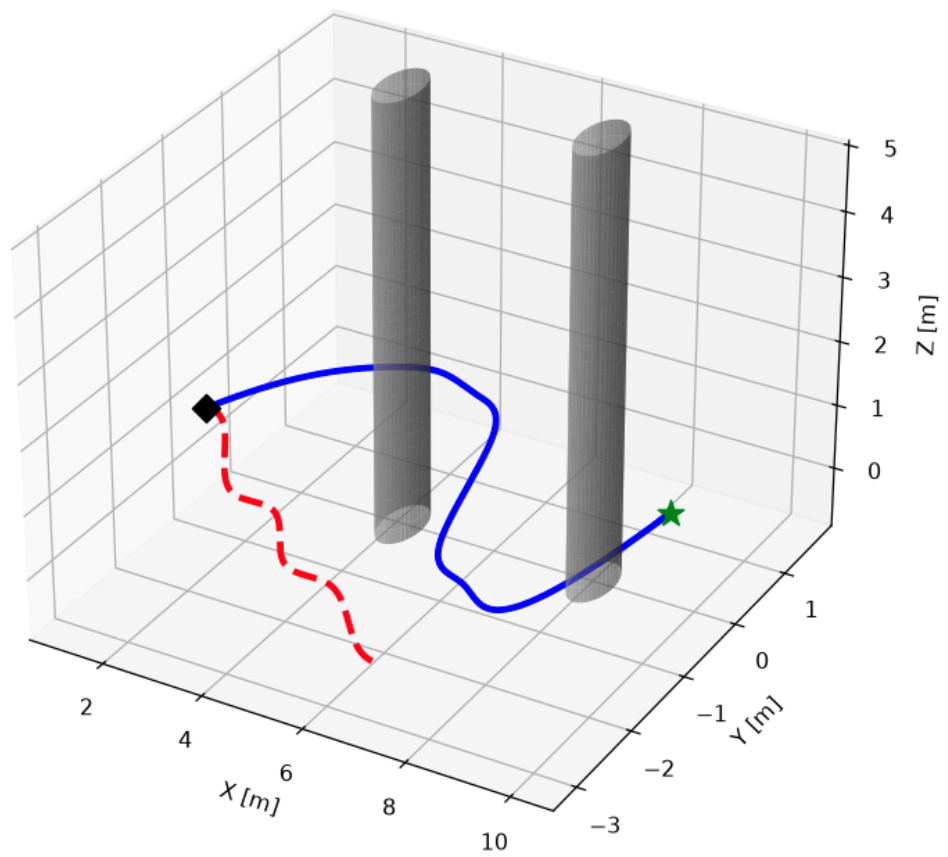}
    \caption{ASV 3-D}
\end{subfigure}
\hfill
\begin{subfigure}[t]{0.46\linewidth}
    \centering
    \includegraphics[width=\linewidth,height=4.2cm,keepaspectratio]
    {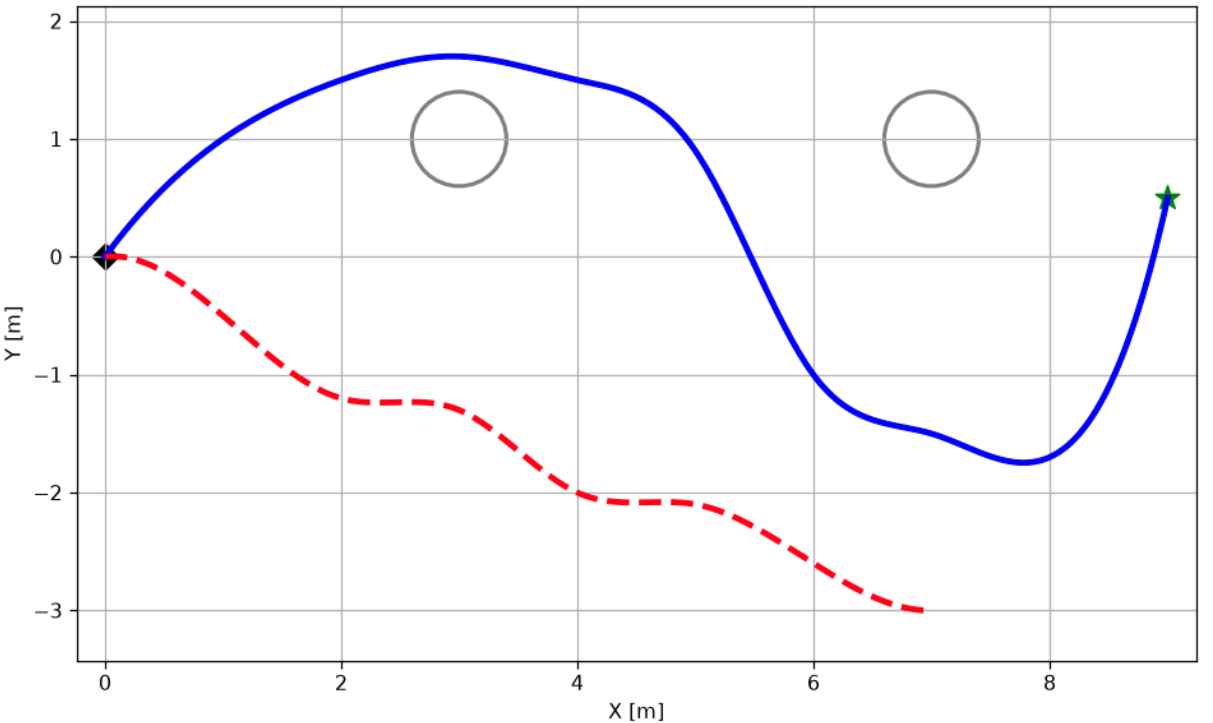}
    \caption{ASV top-down}
\end{subfigure}

\end{minipage}

% \caption{GazeboSim renders and resulting trajectories, one trial of ten per platform (Section~\ref{sec:hifi}). \textbf{(a-c) AUV (\texttt{avoid\_turbine}):} a GazeboSim render near the turbine foundation (a), with the resulting 3-D (b) and top-down (c) trajectories - the world-model planner (blue, solid) detours around the foundation and reaches the goal (green star), while the ungrounded baseline (red, dashed) drifts off course and never arrives. \textbf{(d) Semantic-map input (ASV):} a gridded satellite tile as classified by the VLM pipeline (Section~\ref{sec:vlm_pipeline}), the source of the ASV's obstacle boxes in place of a pre-dive survey. \textbf{(e-g) ASV (\texttt{avoid\_tower}, two turbines):} a GazeboSim render of the scene (e), with the resulting 3-D (f) and top-down (g) trajectories - the planner clears both towers with a turn-forward-turn sequence (no direct lateral detour is possible) and closes on the goal, while the baseline drifts off course. Circles/cylinders mark the true obstacles; the collision field approximates each with an inflated box.}
\caption{GazeboSim renders and resulting trajectories, one trial of ten per platform (Section~\ref{sec:hifi}). \textbf{(a-c) AUV (\texttt{avoid\_turbine}):} render near the turbine foundation (a), with 3-D (b) and top-down (c) trajectories. \textbf{(e-g) ASV (\texttt{avoid\_tower}, two turbines):} render of the scene (e), with 3-D (f) and top-down (g) trajectories. In both, the world-model planner (blue, solid) detours around the obstacle(s) and reaches the goal (green star) - the ASV via a turn-forward-turn sequence, since no direct lateral detour is possible - while the ungrounded baseline (red, dashed) drifts off course and never arrives. \textbf{(d)} Semantic-map input (ASV): a gridded satellite tile as classified by the VLM pipeline (Section~\ref{sec:vlm_pipeline}), the source of the ASV's obstacle boxes in place of a pre-dive survey. Circles/cylinders mark the true obstacles; the collision field approximates each with an inflated box.}

\label{fig:traj_combined}
\end{figure*}

To confirm map-derived obstacles are interchangeable with hand-specified ones, we re-run \texttt{avoid\_tower} and \texttt{slalom} using bounding boxes extracted by the pipeline of Section~\ref{sec:vlm_pipeline} from rendered maps of the same two scenes. \texttt{gemma4:26b} classifies all tower cells correctly (154/154, confidence $>$0.85), and the merged bounding boxes fall within \SI{75}{cm} of the hand-specified ground truth. Planning results are statistically identical to the hand-specified case: 100\% goal-reach, zero collisions, mean goal distance \SI{0.19}{m} (Gradient) and \SI{0.17}{m} (Gradient+MPC) - confirming the pipeline is a drop-in replacement for engineering-drawing obstacle input.

% ============================================================================
\section{Conclusion}
\label{sec:conclusion}

We presented a world-model-grounded LLM planning framework for marine robots near offshore wind infrastructure: the LLM decides \emph{what} to do, the world model decides \emph{for how long}, and an MPC-style loop with a trust-region guard corrects residual drift, on both a 6-DOF AUV and a 3-DOF ASV. The gradient-optimized planner reaches 100\% of goals with zero collisions on both platforms (\SI{0.18}{}-\SI{0.21}{m} AUV, \SI{0.26}{}-\SI{0.28}{m} ASV mean distance) against 0\% goal-reach and dozens of collisions for the ungrounded baseline, and both transfer to GazeboSim after a brief residual fine-tuning pass, cutting goal-distance error by 70-82\% (ASV) and roughly 93\% (AUV), with the same three-phase optimizer and replanning loop applying unchanged across both vehicles.
Two limitations remain open. First, skeleton topology: the optimizer always finds good durations within a fixed skeleton, but cannot repair a wrong topological choice from the LLM. Second, scope: the LLM's role here is deliberately narrow - selecting an ordered macro-action sequence rather than reasoning numerically or semantically - which keeps the symbolic/continuous boundary clean but leaves most of the "reasoning" to the world model and, on the ASV side, to the VLM semantic-mapping stage (Section~\ref{sec:vlm_pipeline}); extending the LLM toward genuine semantic trade-offs (e.g., time vs.\ risk) is a natural next step. The ASV VLM pipeline removes the prior survey assumption, matching hand-specified accuracy on free satellite data. Future work: real-world AUV trials and ASV field deployment.

\bibliographystyle{IEEEtran}
\bibliography{references}

\end{document}